# Cross-Linguistic Offensive Language Detection: BERT-Based Analysis of Bengali, Assamese, & Bodo Conversational Hateful Content from Social Media

Jhuma Kabir Mim*1,*,†*, Mourad Oussalah*2,†* and Akash Singhal*3,†*

*1CVPR , LUT University, Lappeenranta, Finland*
*2CMVS, Faculty of ITEE, University of Oulu, Oulu, Finland*
*3 Dept of Computer Science, University of Helsinki, Helsinki, Finland*

**Abstract**

In today's age, social media reigns as the paramount communication platform, providing individuals with the avenue to express their conjectures, intellectual propositions, and reflections. Unfortunately, this freedom often comes with a downside as it facilitates the widespread proliferation of hate speech and offensive content, leaving a deleterious impact on our world. Thus, it becomes essential to discern and eradicate such offensive material from the realm of social media. This article delves into the comprehensive results and key revelations from the HASOC-2023 offensive language identification result. The primary emphasis is placed on the meticulous detection of hate speech within the linguistic domains of Bengali, Assamese, and Bodo, forming the framework for Task 4: Annihilate Hates. In this work, we used BERT models, including XML-Roberta, L3-cube, IndicBERT, BenglaBERT, and BanglaHateBERT. The research outcomes were promising and showed that XML-Roberta-lagre performed better than monolingual models in most cases. Our team 'TeamBD' achieved rank 3rd for Task 4 - Assamese, & 5th for Bengali.

**Keywords**

Hate Speech, BERT Model, Multilingual Offensive language identification, Social Media, HASOC-2023,

## 1. Introduction

Social media platforms such as Facebook, Twitter, and YouTube stand out as a widely embraced and effortless avenue for uninhibited self-expression and online interaction. Regrettably, it also serves as a platform for disseminating harmful and hostile content, including gender discrimination, xenophobia, protests over politics, online harassment, and even extortion [1]. Hate speech [2] involves derogatory comments targeting an individual or a group due to certain attributes like race, color, or ethnicity. Now this profane language is pervasive on social media, it has become a challenge. Researchers and organizations have been working to develop techniques that can recognize hate speech or abusive language and flag it for review by human moderators or for automated elimination. Consequently, numerous social media platforms actively monitor

---

*Corresponding author.

†These authors contributed equally.

✉ jhuma.mim@student.lut.fi (J. K. Mim); mourad.oussalah@oulu.fi (M. Oussalah); akash.singhal@helsinki.fi (A. Singhal)

user posts. This includes promoting the development of automated techniques for detecting or at least provide insights to detect suspicious and harmful posts. Previous research has explored the identification of offensive language on various platforms, including Twitter [3, 4, 1], Wikipedia comments, and Facebook posts [5], FromSpring posts [6], YouTube [7], and news articles [8], mostly conducted in the English language. The main obstacles in hate speech detection revolve around the lack of essential resources with specific language datasets. Languages with limited resources [9, 10] encounter the difficulty of having inadequately annotated datasets, with only a restricted availability of monolingual datasets. Taking this situation into account, Task 4 of HASOC-23 has the objective of detecting hate speech in Bengali, Bodo, and Assamese languages in 2023 shared task challenge. Advancements [11] in natural language processing (NLP) technology have spurred extensive research into the automated detection of textual hate speech in recent years. Numerous studies [12, 13] have centered their efforts on employing deep learning-based models for the detection of aggressive language within social media text. Academics have also employed FastText embeddings to construct models capable of being trained on billions of words in under ten minutes and classifying millions of sentences into hundreds of categories [14]. There have also been studies [15] indicating how the bias and worldview of annotators can impact the performance of such datasets. Currently, the state-of-the-art research in hate speech detection has advanced to the stage where researchers harness the capabilities of advanced architectures like transfer learning. For instance, in [16], researchers conducted a comparative analysis of deep learning, transfer learning, and multitask learning architectures using an Arabic hate speech dataset. Additionally, another research focused on identifying hate speech within Bengali and hindi datasets through the use of state-of-the-art pretrained models such as XML-Roberta and multilingual BERT(mBERT) [17]. One major hurdle lies in the fact that Bengali is considered a low-resource language. Regrettably, there is a limited amount of research conducted on hate speech detection within Bengali social media resources. Extensive research efforts have been dedicated to developing word embeddings tailored specifically for low-resource languages, with an example being BengFastText [18], designed for Bengali language. In addition, a studied carried out in [19] showed the promise of BanglaHateBERT -a retrained BERT model designed for detecting abusive language in Bengali language. This model underwent training using a substantial corpus of offensive, abusive, and hateful content in Bengali collected from various sources. Likewise, authors of [20] compared various various deep learning models along with pretrained Bengali word embeddings such as Word2Vec, FastText, and BengFastText. Nevertheless, there has been limited research effort conducted in the Assamese and Bodo languages due to low resource. Preliminary works were carried out using transformer architecture (BanglaBERT and mBERT) for abusive language detection in Assamese text [21] & [22], although the authors acknowledged the limited scale and scope of the underlined study, calling for further research efforts. NLP community organized several initiatives to handle this challenge and stimulate research on hateful speech and offensive content in social media, such as Semeval-2019 [23] and 2020 [24]. In the HASOC-2020 [25] & 2021 [26] shared task, it is noteworthy that it drew participation from an extensive pool of over 40 research groups. The attainment of the highest ranking in Hindi hate speech detection was accomplished through the utilization of a Convolutional Neural Network (CNN) incorporating FastText [27] embeddings as input. For the task of German hate speech detection, superior performance was achieved through the deployment of fine-tuned iterations of BERT,

DistilBERT, and RoBERTa [28]. Similarly, the performance in English-language hate speech detection was reached by leveraging BERT alongside another deep learning-based model. These outcomes underscore the diverse and innovative approaches employed by different research groups in addressing the challenge of hate speech detection in varying linguistic contexts. Based on prior research, the BERT model has consistently surpassed the performance of many other contemporary state-of-the-art models. The increasing prominence of BERT stands as a significant trend, underscoring its popularity within the hate speech detection community. In the last five years, BERT has accounted for a substantial share (38%) of deep learning models employed for this purpose [11].

In this year for the first time, 2023 HASOC [29, 30, 31] introduced a novel task involving the identification of multilingual offensive languages for social media, encompassing platforms like Facebook, Twitter, & YouTube. In this contest, our team participated to **Task 4** for Bengali, Bodo, and Assamese languages identification of Hate or offensive. In the context of the state-of-art advancements in hate-speech or offensive text detection, our paper contributes in the following ways:

1. We compare Pre-trained BERT models such as XML-Roberta, L3-cube, IndicBERT, BanglaBERT, and BanglaHateBert for hate speech detection in low resource language.
2. We employed data augmentation using ChatGPT3.
3. We expanded the Dataset through Self-Model Annotation.

The remaining sections of this paper are organized as follows. Section 2 explores the task description and dataset for the three languages. Section 3 details our methodology, encompassing details about the employed transformer models including feature engineering and presents the experimental results, highlighting and discussing the best scoring model. Lastly, conclusive statements and potential future work are drawn in the conclusion section.

## 2. Task Description

Task 4 focuses on identifying hate speech, offensive language, and profanity in different languages using natural language processing techniques. The objective of the task is to detect hate speech in Bengali, Bodo, and Assamese languages. To accomplish this, we used the HASOC-2023 shared dataset for training and validation processes, and cast the problem into a binary classification task [29, 32, 33, 22]. Each dataset (for the three languages) consists of a list of sentences with their corresponding class (hate/offensive (HOF) or non hate/offensive (NOT)).

### 2.1. Datasets

Data is primarily collected from Twitter, Facebook and YouTube comments. The total size of the three datasets all together amounts to 6996 comments, among which 3860 (80%) contain HOF and the rest 3136 (20%) NOT.

1. (NOT) Not-Hate or Offensive - This post does not contain any Hate speech, profane, offensive content.
2. (HOF) Hate or Offensive - This post contains Hate, offensive, and profane content.

**Table 1**
Example of multilingual languages datasets.

| Source | Translation | Label | Language |
|---|---|---|---|
| "বালের শিক্ষা মন্ত্রী" | Stupid Education Minister | HOF | Bengali |
| "তবে শুনলাম মমতা ব্যানার্জি কোটা পদ্ধতি তুলে দিয়েছে তাহলে ব্রাত্য বসুর মুখে কোটা কেন? এটা কথার কথা? তাহলে কোটা নিয়ে প্রাক্তন মুখ্যমন্ত্রীকে কটাক্ষ কেন?" | However, I heard that Mamata Banerjee has given up the quota system, so why is there a quota in Bratya Bose's mouth? Is it word of mouth? So why sneer at the former chief minister about the quota? | NONE | Bengali |
| "দুজেনই মাতাল মাগীবাজ" | Both are drunkards fuckers | HOF | Bengali |
| "কুকুৰ বুলি কিয় কৈছে অসভ্য ক'বৰাৰ, লাজ নাই" | Why are you calling me a dog, rude somewhere, no shame | HOF | Assamese |
| "मोसो खुगायाव एमफौ नांबाय नोंनाव सैम" | - | HOF | Bodo |

## 3. Methodology

This section provides a thorough overview that includes both the model architectures and methodologies used to tackle the task.

### 3.1. Models Descriptions

**BERT** – short for Bidirectional Encoder Representations from Transformers, represents a pioneering language model rooted in transformer architecture [34]. This influential model employs an attention mechanism, thereby enabling the acquisition of contextual relationships among words in a given text sequence. BERT adopts two primary training strategies:

1. *Masked Language Modeling (MLM)* where approximately 15% of the tokens within a sequence are replaced (masked), prompting the model to predict the original tokens.
2. *Next Sentence Prediction (NSP)* where the model is presented with two sentences as input, and it learns to determine whether the second sentence follows the first in their original document context.

**XML-Roberta** – XML-Roberta is a variant of the RoBERTa model, which is a popular transformer-based language model because of its ability to handle cross-lingual tasks. It has been fine-tuned to perform exceptionally well in various languages and is particularly robust in multilingual settings. The "XML" refers to "Cross-lingual Multilingual," emphasizing its proficiency in understanding and generating text across more than 100 languages [35] at once.

**IndicBERT** – IndicBERT represents a multilingual ALBERT model that has been exclusively pretrained on a comprehensive dataset comprising 12 major Indian languages [36]. Following pretraining, IndicBERT underwent evaluation across a range of diverse natural language understanding tasks.

**L3-cube** – It is a mBERT model fine-tuned on L3Cube-HingCorpus. The latter is the first large-scale real Hindi-English code mixed data known as HingBERT [37] in a Roman script.

**BanglaBERT** – Bangla-Bert-Base [38] is a pre-trained Bengali language model that employs the mask language modeling technique, as outlined in the BERT framework.

**BanglaHateBERT**– BanglaHateBERT [19] model is essentially designed for abusive and offensive language identification in Bengali.

### 3.2. Experiments and Results

In the context of Task 4, we embraced three primary steps- based methodology:

**Exploration stage:** We conducted experiments with a range of BERT models, both multilingual (e.g., XML-Roberta-large and IndicBERT) and monolingual (e.g., L3-cube, Bangla BERT, and Bangla Hate BERT) [1]. We have used in each case 90% od dataset for training and 10% for the testing and validation. After conducting our experiments, we decided to choose the XML-Roberta-large model as the baseline model because it performed better than all other monolingual models. The performance disparity observed between certain monolingual models, such as IndicBERT or Bangla Hate BERT, when compared to XML-Roberta-large, can potentially be attributed to the fact that these models are not consistently up-to-date. Besides, observing that XML-Roberta-large consistently beats three monolingual models, we decided to use it as the default model. Table 2 presents F1 scores for different models and methods by Language.

**Data Augmentation Process:** In our study, we employed data augmentation technique to enhance the robustness and diversity of our dataset. We utilized ChatGPT-3, a state-of-the-art language model to perform data augmentation, ensuring that our text samples maintained their original annotation labels, specifically distinguishing between "HOF" (Hate Speech or Offensive) and "Not HOF" (Non-Hate Speech or Non-Offensive). Our approach aimed to generate additional samples while preserving the integrity of the original labels. To conduct this operation, we selected ChatGPT-3 as our data augmentation tool due to its impressive language generation capabilities and its ability to produce coherent and contextually relevant text. We formulated a specific prompt for ChatGPT-3 to guide its generation of augmented samples. The prompt played a critical role in instructing the model to maintain the original annotation label while generating new content. An example of our prompt is as follows:

"' Given the following text sample: [Original Text Sample], please generate three additional samples that preserve the original annotation label (HOF or NOT). "'

This prompt structure ensured that ChatGPT-3 produces three augmented versions of each input text while respecting the initial annotation. To assess the effectiveness of our data augmentation process, we manually evaluated a random sample of 200 augmented results generated by ChatGPT-3. During this evaluation, we compared the original annotation labels with those of the augmented samples. We found that a remarkable 98% of the augmented

---
[1]https://github.com/Meem007/Hate-Speech-and-Offensive-Content-Indentification

**Table 2**

F1 Scores for Different Models by Language of Task 4

| Language | Model | F1 Score |
|---|---|---|
| Assamese | **xlm-roberta-large (multilingual) + ChatGPT3 augmentation** | **72.2** (submitted) |
| | xlm-roberta-large (multilingual) + 5k self annotated data | 73.3 |
| | xlm-roberta-large (multilingual), before data augmeneation | 71.2 |
| | IndicBERT (Pre-trained on 12 major Indian languages) | 69.1 |
| | BanglaBERT (monolingual) | 65.1 |
| | BanglaHateBERT (monolingual) | 61.5 |
| | L3-cube (monolingual) | 60.5 |
| Bengali | **xlm-roberta-large (multiingual)** | **73.4** (submitted) |
| | IndicBERT (Pre-trained on 12 major Indian languages) | 70.5 |
| | BanglaBERT (monolingual) | 65.5 |
| | BanglaHateBERT (monolingual) | 68.1 |
| | L3-cube (monolingual) | 64.4 |
| Bodo | **xlm-roberta-large (multiingual)** | **76.2**(submitted) |
| | IndicBERT (Pre-trained on 12 major Indian languages) | 70.5 |
| | BanglaBERT (monolingual) | 50.1 |
| | BanglaHateBERT (monolingual) | 49.1 |
| | L3-cube (monolingual) | 55.5 |

samples accurately preserve the original label. Table 3 presents example sentences generated by ChatGPT3.

**Expanding the Dataset Through Self-Model Annotation:**

To further bolster our model's performance, we targeted exclusively Bengali language due to the unavailability of suitable datasets for Assamese and Bodo. For this purpose, we implemented a complementary strategy aimed at enlarging our training dataset by incorporating additional data. We leveraged an additional Bengali public dataset containing 30,000 samples from a different domain, which we annotated using our initial baseline model [2].

The approach involved the initially training the model using 90% of the existing training data and reserving 10% for testing & evaluation purposes. Subsequently, during the evaluation phase, we established optimal thresholds and applied both upper and lower thresholds to automatically label a portion of the test data. For example, we utilized an upper threshold of 0.90 and a lower threshold of 0.20. Following the automatic annotation of this portion of the new public data using these thresholds, we retrained the model by incorporating this annotated data into the training dataset. Remarkably, this process led to a 2.1% improvement in the model's performance.

Our initial model employed a threshold of 0.50 for classification decisions. Recognizing the potential for improved results by varying these thresholds, we embarked on an empirical journey to explore higher and lower threshold values. It is important to emphasize that our threshold selection process was not grounded in established scientific methodologies but rather conducted as an experiment guided by intuition. We acknowledged that the chosen threshold values lacked rigorous scientific validation. Nonetheless, the results we obtained were empirically favorable.

---

[2]Source: https://www.kaggle.com/datasets/naurosromim/bengali-hate-speech-dataset

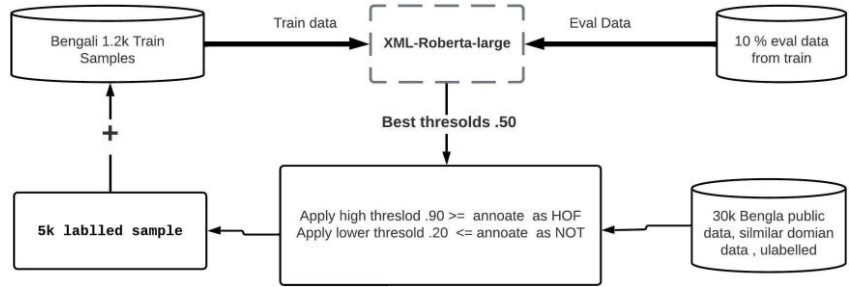

**Figure 1:** Dataset Expansion Through Self-Model Annotation

With this annotation technique, we were able to label an additional 5,000 samples out of the 30,000 in the public Kaggle dataset. However, it is important to note that, due to shared task guideline, we did not include these results as part of our final submission.

**Table 3**

Example of ChatGPt data augmentation for Bengali

| Source | Translation | Label | Language |
|---|---|---|---|
| **Original Sentence** "বালের শিক্ষা মন্ত্রী" | Stupid Education Minister | HOF | Bengali |
| **Augmented Sentences** "বালের শিক্ষা মন্ত্রী খুব বেশি মন্ত্রণা করে।" | The Stupid Education Minister for children does a lot of counseling. | HOF | Bengali |
| "বালের শিক্ষা মন্ত্রী শিক্ষার ক্ষেত্রে কাজ করে।" | The Stupid Education Minister for children works in the field of education. | HOF | Bengali |
| "বালের শিক্ষা মন্ত্রী প্রশাসনিক কাজে নিয়োজিত।" | The Stupid Education Minister for children is appointed for administrative work. | HOF | Bengali |

**Table 4**

Training Size Before and After Data Augmentation. We have augmented only the training samples while keeping the validation samples untouched.

| Language | Training Size (Before) | Training Size (APer Augmentation) |
|---|---|---|
| Bengali | 1,281 | 3,843 |
| Assamese (Ashamee) | 4,036 | 12,108 |
| Bodo | 1,679 | 5,037 |

**Table 5**
Official results of our HASOC-23 test set submissions for Task 4 aims to detect hate speech in Assamese Bengali, and Bodo languages.

| Team name | Task, Language | Macro F1 | Rank |
|---|---|---|---|
| TeamBD | Task 4 (Assamese) | 0.722 | 3 |
| TeamBD | Task 4 (Bengali) | 0.734 | 5 |
| TeamBD | Task 4 (Bodo) | 0.762 | 14 |

## 4. Conclusion

In addition to demonstrating the efficiency of the transformer-based models in identifying abusive language, our research has illuminated the significance of model choice, task-specific modifications, and creative approaches needed to tackle the challenges posed by multilingual and cross-lingual scenarios. Our primary focus was on Task 4, which involves detecting offensive language in Assamese, Bengali, and Bodo languages. To handle the inherent complexities of these tasks, we leveraged powerful transformer-based models, including XML-Roberta, among others.

Through rigorous assessments, a consistent trend emerged: XML-Roberta-large is found to consistently outperform monolingual models across the majority of test scenarios. This observation underscores the efficacy of employing cross-lingual pre-trained models, particularly in resource-constrained settings.

Furthermore, our data augmentation strategy, enabled by ChatGPT-3, has proven to be remarkably successful in enlarging our dataset's size and diversity, while preserving the reliability of the original annotation labels. Notably, we observed substantial improvements of nearly 1.2-1.5% across almost all three languages after augmentation. As part of future work, it would be intriguing to compare this approach with other large language models (LLMs) and existing traditional methods of text augmentation.

To enhance the performance of our model, we expanded our training dataset exclusively for Bengali using Self-Model annotation. We annotated this data with the help of our baseline model and conducted experiments with various classification thresholds. Despite the lack of scientific validation for threshold selection, this approach resulted in a noteworthy 2.1% performance improvement. This technique shows a promise for domain adaptation and bolstering model robustness. It is intriguing that, even though the model already had knowledge of this additional data, however, adding self-annotated data further enhanced its learning, highlighting the feasibility and potential benefits of such experiments.

Our top-ranking results in Task 4 for Bengali and Assamese underscore the versatility and reliability of these models in addressing real-world challenges. In conclusion, our research may make a substantial contribution to the formidable task of identifying objectionable language in multilingual and cross-lingual contexts. We anticipate that our findings will inspire further scholarly investigations in this field, fostering the development of more potent and reliable methods for detecting and mitigating offensive language in online discourse.

## 5. Acknowledgments

We want to convey our appreciation to the HASOC-2023 organizers for affording us the chance to engage in this shared task and for their assistance throughout the duration of the event. Furthermore, we recognize the valuable contributions of our entire research team and the resources made available by the NLP community, which were instrumental in facilitating this research.